\documentclass{l4dc2025}

\usepackage{bm}
\usepackage{algorithm}
\usepackage{algorithmic}
\usepackage{footnote}
\newtheorem{assumption}{Assumption}

\title[Black-box Environment Poisoning]{Online Poisoning Attack Against Reinforcement Learning under Black-box Environments}
\usepackage{times}
\author{%
 \Name{Jianhui Li} \Email{jianhuili@zju.edu.cn}\\
 \addr College of Control Science and Engineering, Zhejiang University, P. R. China\\
  \addr School of Data Science, The Chinese University of Hong Kong, Shenzhen, P. R. China
 \AND
 \Name{Bokang Zhang} \Email{bokangzhang@link.cuhk.edu.cn}\\
 \addr School of Science and Engineering, Chinese University of Hong Kong, Shenzhen, P. R. China
 \AND
 \Name{Junfeng Wu} \Email{junfengwu@cuhk.edu.cn}\\
 \addr School of Data Science, Chinese University of Hong Kong, Shenzhen, P. R. China
}

\begin{document}

\maketitle

\begin{abstract}%
This paper proposes an online environment poisoning algorithm tailored for reinforcement learning agents operating in a black-box setting, where an adversary deliberately manipulates training data to lead the agent toward a mischievous policy. In contrast to prior studies that primarily investigate white-box settings, we focus on a scenario characterized by \textit{unknown} environment dynamics to the attacker and a \textit{flexible}  reinforcement learning algorithm employed by the targeted agent.
 We first propose an attack scheme that is capable of poisoning the reward functions and state transitions. The poisoning task is formalized as a constrained optimization problem, following the framework of \cite{ma2019policy}. Given the transition probabilities are unknown to the attacker in a black-box environment, we apply a stochastic gradient descent algorithm, where the exact gradients are approximated using sample-based estimates. A penalty-based method along with a bilevel reformulation is then employed to transform the problem into an unconstrained counterpart and to circumvent the double-sampling issue. The algorithm's effectiveness is validated through a maze environment.  
\end{abstract}


\section{Introduction}
\label{sec:introduction}
Reinforcement learning (RL) algorithms have gained substantial popularity in recent years, finding applications across a variety of fields such as robotic manipulation (\cite{gu2016deep}), recommendation systems (\cite{chen2019generative}), and autonomous vehicles (\cite{o2018scalable}). Despite their success, these algorithms typically depend on vast amounts of collected data for effective agent training. This reliance presents a significant vulnerability: adversaries can disrupt the training process by injecting poisoned data, thereby steering the agent towards adopting potentially malicious policies. A striking example is Tay, a conversational AI chatbot developed by Microsoft that incorporates adaptive learning capabilities. Through the coordinated efforts of certain malicious Twitter users, Tay was manipulated into producing racist and antisocial content (\cite{neff2016talking}).  \par

The paper \cite{kiourti2020trojdrl} was one of the pioneering works to investigate the design of data poisoning attacks against reinforcement learning (RL) agents. It introduced a man-in-the-middle attack (MIMA) framework that alters state data. In this framework, the adversary intercepts and manipulates the data from the environment to the agent, effectively directing the agent to adopt a predefined target policy. Subsequent studies have expanded this line of research to encompass scenarios with limited attack budgets \cite{rakhsha2021reward, zhang2020adaptive}, sparse attacks with stealthy awareness~\cite{cui2024badrl}, and multi-agent environments~\cite{wu2023reward, liu2023efficient}. 
However, a significant limitation of the majority of these approaches is their assumption of prior knowledge about reward and transition models.
This assumption is less realistic in the context of RL, as RL typically operates in an online setting. \par 

In this paper, we focus on an online MIMA in a black-box context, wherein the adversary operates without prior knowledge of environment dynamics. The attacker is capable of modifying the reward and transitioned state data. To enhance the stealthiness of attack, we formalize the task as an optimization problem aimed at minimizing the deviation from the original MDP. Furthermore, the modified transitioned states are restricted to those that are reachable in the original environment. This attack is termed \textit{online} because it takes place simultaneously with the agent-environment interaction over time. Our main contributions are summarized as follows:
\begin{enumerate}
    \item In contrast to the previous transition probability poisoning algorithms that primarily focus on white-box settings (such as \cite{rakhsha2020policy}), we propose a more practical poisoning 
    framework that operates effectively in scenarios where the environment dynamics are unknown to the attacker and the targeted agent employs a flexible reinforcement learning algorithm.
    
    \item  We develop an iterative optimization algorithm for seeking an optimal poisoning tactics. This method employs a penalty-based approach combined with a bilevel reformulation to avoid the so-called double-sampling issue. In addition, we leverage stochastic gradient descent technique in bilevel optimization, 
    enabling an efficient online poisoning execution. 
\end{enumerate}
\textbf{Notation}: The indicator function $\mathbf{1}(\cdot)$ outputs $1$ if the event of interest is true and $0$ otherwise. The cardinality of a finite set $\mathcal{A}$ is denoted as $|\mathcal{A}|$. To differentiate between variables related to the attacker and those related to the RL agent, we use overlined symbols to represent the attacker's variables.

\subsection{Related Works}
Existing research on data poisoning attacks against RL agents have been studied under different contexts. 

\textbf{Black-box vs. White-box Environments}. In white-box scenario, attackers are usually assumed to have prior knowledge of the environment dynamics, while attackers in black-box settings posit that such information is unavailable to them. 
Most existing studies (\cite{kiourti2020trojdrl, ma2019policy, rakhsha2020policy, foley2022execute}) focus primarily on white-box scenarios, which may be practical for supervised learning victims but are less realistic in the context of reinforcement learning. The paper \cite{rakhsha2021reward} investigates an online attack, in which the environment dynamics is unknown. The authors categorize the attack into two phases: the exploration phase and the attack phase. However, establishing the transition criteria between the two phases poses a significant challenge.  The work \cite{xu2023black} describes an approach where the attacker adjusts the rewards of actions based on their proximity to the target action. Both of these methods employ heuristic strategies.  %our approach adopts an optimization-based method that may reduce the deviations from the original environment dynamics. \par

\textbf{Poisoning Context}. 
In the realm of adversarial attacks, attackers are assumed to be capable of poisoning various components of data.
Research has explored the manipulation of state information~\cite{ashcraft2021poisoning} and action poisoning~\cite{liu2021provably}.  However, a significant number of studies concentrate on altering reward data (\cite{ma2019policy, zhang2020adaptive, wu2023reward, rangi2022understanding}), as rewards are typically manually designed and tend to be less sensitive to minor changes. 
Some studies also explore poisoning of both reward and transition probabilities (\cite{rakhsha2020policy, xu2022spiking}). Notably, \cite{rakhsha2020policy} examines a white-box attack in which transition probabilities are assumed to be known.  \cite{xu2022spiking} presents a method for poisoning both the reward data and transition probabilities in a black-box environment setting, but fails to provide a viable scheme for modifying environment's hyper-parameters to alter transition dynamics.

\section{Preliminary and Problem Formulation}
\label{sec:preliminary}
\begin{figure}[t]
    \centering
    \includegraphics[width=0.7\linewidth]{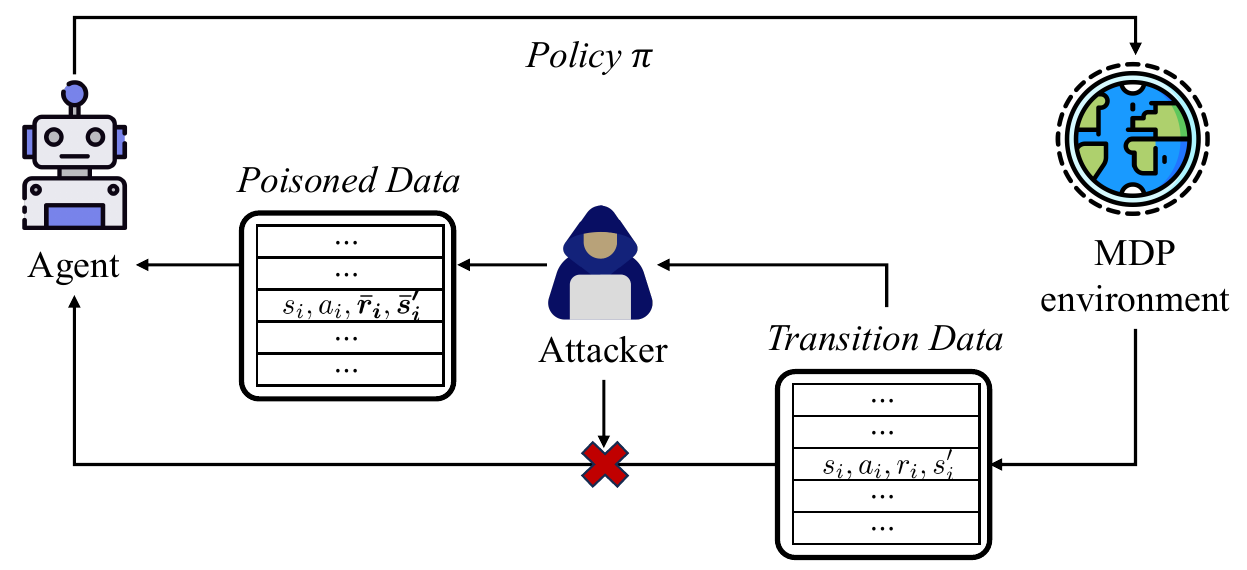}
    \caption{The interaction involves three entities: the agent, the environment and the attacker. The agent communicates its policy $\pi$ to the environment, which then generates transition data based on the policy. However, the attacker intercepts the process, manipulating the rewards and the transitioned states in the data. The poisoned data is then fed back to the agent and is used to update the agent's policy. In the figure, italicized text represents data, while non-italicized text refers to entities. }\vspace{-15pt}
    \label{fig:poisoning}
\end{figure}

In this section, we introduce three key entities in the RL data poisoning problem: the Markov decision process (MDP) that describes the environment, the RL agent, and the attacker that utilizes poisoning action. Their interactions are illustrated in Figure \ref{fig:poisoning}.  \par
\textbf{MDP Environment}. We model the environment as a discounted MDP characterized by a tuple $\langle\mathcal{S}, \mathcal{A}, r, P, \gamma \rangle$, where $\mathcal{S}$ denotes a finite state space, and $\mathcal{A}$ denotes a finite action space. The reward $r\in\mathbb{R}^{|\mathcal{S}||\mathcal{A}|}$ is specified as $r=[r_{s,a}]_{s\in\mathcal{S},a\in\mathcal{A}}$.  The transition probability is defined as $P:\mathcal{S}\times\mathcal{A}\times\mathcal{S}\rightarrow [0,1]$. The discount factor $\gamma\in(0,1]$ reflects our preference for future rewards against immediate ones. \par

\textbf{RL Agent}. The RL agent interacts with the environment, collecting transition data in the form of $\langle s,a,r_{s,a},s'\rangle$ over time, where $s'$ is the transitioned state that results from transitioning from  $s$,  following the distribution $P(\cdot|s,a)$, when action $a$ is applied. The agent's objective is to maximize the discounted accumulated reward by acting according to a policy $\pi:\mathcal{S}\times\mathcal{A}\rightarrow [0,1]$, where reward is calculated in expectation. The $Q$ function\footnote{While an RL agent may employ an arbitrary RL algorithm, including a non-Q-learning algorithm, for its training, we can still define the $Q$ function in this context
}, which quantifies the expected cumulative reward for state $s$ and action $a$ derived from policy $\pi$, is defined as
\begin{equation}\label{eqn:Q_function}
Q^\pi_{s,a}:=\mathbb{E}_{a_t\sim\pi(\cdot|s_t), s_{t+1}\sim P(\cdot|s_t, a_t)}\Big[ \sum_{t=0}^\infty \gamma^t r_{s_t,a_t}|s_0=s, a_0=a \Big].
\end{equation}
The Bellman equation of $Q$ function can be derived as follows:
\begin{equation}\label{equ:bellman_equation}
    Q^\pi_{s,a} = r_{s,a}+ \gamma \sum_{s', a'} P(s'|s,a)\pi(a'|s') Q^\pi_{s',a'} .
\end{equation}

% \bk{More like Threat Model? or Problem Definition}. \jh{Corresponds to the first sentence in this subsection, that introduce MDP, agent, and attacker sequentially. }

\textbf{RL Poisoning Attacker}. The attacker sits between the environment and the agent, with the capability of manipulating both the reward $r$ and the transitioned state $s'$ of the transition data collected by the RL agent.  By replacing the original data record $\langle s,a,r_{s,a},s' \rangle$ with $\langle s,a,\bar{r}_{s,a}, \bar{s}' \rangle$, where $\bar{r}$ is the poisoned reward,  $\bar{s}'$ is the poisoned state, the compromised record $\langle s,a,\bar{r}_{s,a}, \bar{s}' \rangle$ is then utilized  by the agent for RL training. It is important to note that the attacker does not alter the transition probability $P$ of the MDP environment directly. Instead, it manipulates the state transition trajectory, as the RL algorithms of interest are sampling-based.  We further assume that the probabilistic transition model is a black box to the agent providing the attacker with an opportunity to remain stealthy and evade detection. Finally, the attacker does not need to know the specific reinforcement learning algorithm that the agent uses for training, further complicating the detection and mitigation of the attack. 
To carry out data poisoning effectively, we require the attacker to satisfy the following assumption: 
\begin{assumption}
The attacker has knowledge of the reachable set,
that is, the set of all possible state transitions from state $s$ under action $a$, defined as $\mathcal{S}'_{s,a}:=\{ s'\in\mathcal{S}:P(s'|s,a)>0 \}$, for all $s\in\mathcal{S}$ and $a\in\mathcal{A}$.
\end{assumption}
This assumption is considered mild, as such information can be acquired through agent-environment interactions prior to data poisoning or learned online in parallel with the poisoning attack. More sophisticated studies on this topic will be reserved for future work.\par

In this paper, we are interested in a policy induction attack goal, wherein the attacker intends to steer the agent towards learning a pre-defined target policy $\bar \pi$ of its own design. Following  \cite{ma2019policy}, the target policy is restricted to be deterministic, i.e., $\bar \pi:\mathcal{S}\rightarrow\mathcal{A}$. This restriction aligns with the behavior observed in most reinforcement learning algorithms. Although a stochastic policy may be employed during the training, it ultimately converges to a deterministic policy which choose certain action with probability $1$.

\section{Main Results}
\label{sec:proposed_method}
In this section, our poisoning attack algorithm is presented. Section \ref{subsec:attack_scheme} {introduces a practical method to alter the transition probabilities under black-box setting through manipulating transitioned states}. Section \ref{subsec:reformulation} formalizes the poisoning activities as an optimization problem through a penalty-based approach. Furthermore, a bilevel formulation is taken to resolve the double-sampling issue when samples from 
environment-agent interaction are used to approximate the transition probabilities in the black-box environment.
The bilevel formulation, though complicates the
problem to some extent, effectively avoids the computation of problematic gradients. Finally, a single-loop stochastic gradient descent method for bilevel optimization is derived in Section \ref{subsec:algorithm}  having successfully addressed the computation of approximated gradients.

%\jh{The subsequent roadmap is outlined as follows. Given unknown transition probabilities, attacker has to use samples to approximate the transition probabilities. We first transform the constrained optimization problem to an unconstrained optimization problem through a penalty-based approach. This conversion, however, introduces a double-sampling issue associated with the equality constraints. Therefore, we further adopt a bilevel reformulation, which though complicates the problem, avoids the computation of problematic gradients. }\jf{need to mention penalty method!! }

\subsection{Manipulating Transitioned State}
\label{subsec:attack_scheme}
As discussed in Section \ref{sec:preliminary}, the attacker is capable of altering the reward data and the transitioned state.  The modified training data will be fed into the agent  to steer it towards $\bar\pi$ via shifting the equilibrium of the Bellman equation~\eqref{equ:bellman_equation}.
This is achieved by influencing the reward function $r_{s,a}$ through poisoned rewards and the transition probability $P(s'|s,a)$ via manipulated transitioned states. Adjusting  rewards seems straightforward since \eqref{equ:bellman_equation} is an explicitly linear function of $r_{s,a}$, while modifying $s'$ is challenging since 
$P(s'|s,a)$ is not directly accessible to the attacker.  In what follows, we will introduce a scheme that enables the attacker to decide on  $s'$ leveraging it knowledge of the reachable set  $\mathcal{S}'_{s,a}$ for
state $s$ under action $a$.
In particular, when the transition from $s$ to $s'$ is observed in the transition data given action $a$,  the attacker randomly  draws a
state $\bar s'$ from $\mathcal{S}'_{s,a}$ according to a uniform distribution. Subsequently,  it replaces $s'$ with $\bar s'$ with a probability $\delta_{s,a}\in [0,1]$ while leaves $s'$ unchanged with probability $1-\delta_{s,a}$.
This scheme leads to the following modified transition probability
\begin{equation}\label{equ:poisoned_transition}
    \bar{P}_\delta(s'|s,a) = (1-\delta_{s,a})P(s'|s,a) + \delta_{s,a} \frac{\mathbf{1}(s'\in\mathcal{S}_{s,a}')}{|\mathcal{S}_{s,a}'|}.
\end{equation}
It unveils that the modified transition probability $\bar{P}_\delta(s'|s,a)$ is a convex combination of $P(s'|s,a)$ and a uniform distribution on $\mathcal S'_{s,a}$, controlled by the parameter $\delta_{s,a}$. However,  this modification
is restricted to one-dimensional adjustments, owing to the limited representational capability of two-component convex combinations. \par

%However, the dynamics-agnostic nature complicates the manipulation of transition probabilities, which is illustrated in the following example. 
%\begin{example}
%Consider a scenario where a particular state-action pair %$s,a$ leads to two potential transitioned states, namely $A$ and $ B$, with respective transition probabilities of 0.8 and 0.2. To modify the distribution to $0.6$ for state A and $0.4$ for state B, the attacker could replace each instance of $A$ with $B$ for $25\%$ probability. In expectation, this modification would yield the target distribution $(0.6, 0.4)$. However, determining the probability of substitution $25\%$, requires knowing the original transition probabilities.
%\end{example}

%Therefore, we propose an alternative attack scheme where the original transitioned state is substituted with a uniformly sampled state from the set of reachable transitioned state denoted as $\mathcal{S}'_{s,a}:=\{ s'\in\mathcal{S}:P(s'|s,a)>0 \}$. Specifically, with probability $\delta_{s,a}\in[0,1]$, the attacker substitutes the original transitioned state $s'$ with $\bar{s}'$, a state selected uniformly from $\mathcal{S}_{s,a}'$. The attack scheme leads to a modified transition probability

%\begin{equation}\label{equ:poisoned_transition}
   % \bar{P}_\delta(s'|s,a) := (1-\delta_{s,a})P(s'|s,a) + \delta_{s,a} \frac{\mathbf{1}(s'\in\mathcal{S}_{s,a}')}{|\mathcal{S}_{s,a}'|}.
%\end{equation}

In a nutshell, the attacker's poisoning attack is characterized by the poisoned reward vector $\bar{r}\in\mathbb{R}^{|\mathcal{S}||\mathcal{A}|}$, and a vector  $\delta\in[0,1]^{|\mathcal{S}||\mathcal{A}|}$ of the parameters $\delta_{s,a}$ reflecting the intensity of poisoning transitioned states. \par

\subsection{Formulating Poisoning as an Optimization Problem}
\label{subsec:reformulation}

Based on the characterization of the poisoning activity at the attacker side,  the attacking task can be formalized as follows:
%Building on the framework of \cite{ma2019policy},
\begin{equation}\label{equ:org_formulation}
    \begin{aligned}
        \mathop{\text{minimize}}_{\bar{r}, \delta, \bar{Q}} \quad & \frac{1}{2}\sum_{s,a}( \bar{r}_{s,a} - r_{s,a} )^2 + \frac{\rho_\delta}{2}\sum_{s,a}\delta_{s,a}^2 \\
        {\text{subject~to}} \quad & \bar{Q}_{s,a} = \bar{r}_{s,a} + \gamma \sum_{s'}\bar{P}_\delta(s'|s,a)\bar{Q}_{s',\bar \pi_{s'}}, \quad \forall s,a\\
        & \bar{Q}_{s,\bar \pi_s} \geq \bar{Q}_{s,a} + \epsilon, \quad \forall s, a\neq \bar \pi_{s},
    \end{aligned}
\end{equation}
where $\rho_\delta > 0$ is a weight  that controls the attacker's flavor on transition poisoning  relative to reward poisoning, $\bar{Q}\in\mathbb{R}^{|\mathcal{S}||\mathcal{A}|}$ denotes the $Q$-value that the attacker aims to induce, and $\bar{r}\in\mathbb{R}^{|\mathcal{S}||\mathcal{A}|}, \delta\in\mathbb{R}^{|\mathcal{S}||\mathcal{A}|}$ are the decision variables for poisoning. The equality constraints correspond to the shifted Bellman equation~\eqref{equ:bellman_equation}, and the inequality constraints ensure a value gap of $\epsilon>0$ between the $Q$-values of target actions and those of the other actions. Note the distinction of  $\bar{Q}$, the variables controlled by the attacker, and $Q$-value defined in~\eqref{eqn:Q_function}. \par%Nonetheless, with a successful attack and provided that the agent is properly trained, the $Q$-value function estimated by the agent will ultimately converge to $\bar{Q}$. \par

The optimization problem~\eqref{equ:org_formulation} cannot be solved directly as $P(s'|s,a)$ in~\eqref{equ:poisoned_transition} is not known a priori at the attacker. Consequently, it is necessary to approximate this probability using transition data record sampled from the environment by the agent.  To facilitate the resolution of ~\eqref{equ:org_formulation}, we will  further explore the possibility of shifting constraints to the objective function. 
%Therefore, we employ a stochastic gradient descent method, necessitating the transformation of the current constrained optimization problem~\eqref{equ:org_formulation} into an unconstrained optimization problem. 
The Lagrangian and penalty methods are two well-established approaches for such conversion. Lagrangian methods involve introducing multipliers. This may complicate the problem, particularly when there are a large number of constraints, as is the case with ~\eqref{equ:org_formulation}. Penalty methods avoid the need for additional variables by incorporating penalty functions as an softened approximation of the constraints. In this work, we adopt a penalty-based approach. \par

\textbf{Penalty-based method}. We first convert the inequality constraints to penalty functions and shifted to the objective function. The optimization problem~\eqref{equ:org_formulation} is reformulated as follows, 
\begin{equation}\label{equ:formulation_penalty}
    \begin{aligned}
\mathop{\text{minimize}}_{ \bar{r},\delta, \bar{Q}} \quad & \frac{1}{2}\sum_{s,a}(\bar{r}_{s,a} - r_{s,a})^2 +\frac{\rho_\delta}{2} \sum_{s,a}\delta_{s,a}^2+\frac{\rho_\Phi}{2} \sum_{s,a\neq\bar{\pi}_s} (\Phi(\bar{Q}_{s,a}+\epsilon-\bar{Q}_{s,\bar \pi_s}))^2\\
        \text{subject~to}  \quad & \bar{Q}_{s,a} = \bar{r}_{s,a} + \gamma\sum_{s'}\bar{P}_\delta(s'|s,a)\bar{Q}_{s',\bar \pi_{s'}} ,\quad \forall s, a,
    \end{aligned}
\end{equation}
where $\rho_\Phi>0$ is the penalty coefficient, $\Phi(\cdot)$ is defined as $\Phi(x):=\mathbf{1}(x>0)x$. To guarantee the equivalence between the solution of~\eqref{equ:formulation_penalty} and that of~\eqref{equ:org_formulation}, an iteratively increasing $\rho_{\Phi,k}$ is employed following the convention of penalty-based method (\cite{nedic2020convergence}). This approach ensures that the penalty function becomes dominant at sufficiently large iterations. \par

Ideally, both the inequality and equality constraints could be converted into penalty functions. However, penalty functions induced from the equalities lead to the double sampling issue (\cite{dai2018sbeed}). Specifically, the penalty would involve the term $(\bar{Q}_{s,a}-\bar{r}_{s,a}-\gamma\sum_{s'}\bar{P}_\delta(s'|s,a)\bar{Q}_{s',\bar \pi_{s'}})^2$ from the equality constraints. Differentiating this term with respect to $\bar{Q}_{\tilde s,\tilde a}$ for some $\tilde s\in\mathcal{S}, \tilde a\in\mathcal{A}$, we obtain that
\begin{equation*}
\begin{aligned}
    & \nabla_{\bar{Q}_{\tilde s,\tilde a}} (\bar{Q}_{s,a}-\bar{r}_{s,a}-\gamma\sum_{s'}\bar{P}_\delta(s'|s,a)\bar{Q}_{s',\bar \pi_{s'}})^2 \\
    = & (\bar{Q}_{s,a}-\bar{r}_{s,a}-\gamma\sum_{s'}\bar{P}_\delta(s'|s,a)\bar{Q}_{s',\bar \pi_{s'}}) (\mathbf{1}(s=\tilde s, a=\tilde a) - \gamma \mathbf{1}(\tilde a=\bar \pi_{\tilde s})\bar{P}_\delta(\tilde s|s,a)).
\end{aligned}
\end{equation*}
The gradient pertains a probability product  $\bar{P}_\delta(s'|s,a)\bar{P}_\delta(\tilde{s}|s,a)$,  
necessitating two independent samples to derive an unbiased estimate of the gradient
for implementing a first-order optimization algorithm. In most model-free RL applications, a simulator oracle capable of repeated state propagation is often unrealistic. Therefore, we refrain from integrating the equality constraints into the objective. 

The paper by \cite{dai2018sbeed} addresses an optimization similar~\eqref{equ:formulation_penalty}  using a primal-dual method. Nonetheless, as a Lagrangian approach, their method can be less efficient when dealing with large sets of actions or states, as it requires the introduction of numerous dual variables. As an alternative, we propose using a bilevel optimization method to tackle this challenge.\par

\textbf{Bilevel Optimization}. Bilevel optimization encompasses a hierarchical structure of two nested optimization problems, namely the upper- and lower-level problems, which are interdependent. To transform the bilevel optimization problem from~\eqref{equ:formulation_penalty}, , we must interpret the equalities in~\eqref{equ:formulation_penalty} as the KKT condition of the low-level optimization problem. Specifically, the condition
$$\bar{Q}_{s,a} = \bar{r}_{s,a} + \gamma\sum_{s'}\bar{P}_\delta(s'|s,a)\bar{Q}_{s',\bar \pi_{s'}}$$
holds for all $s\in\mathcal S$ and $a\in\mathcal A$
if and only if
$\bar{r} = \mathop{\arg\min}_{\bar{r}}
g(\bar{r},\delta, \bar{Q})$, where
$$g(\bar{r},\delta, \bar{Q}):= \sum_{s,a}  \frac{1}{2}(\bar{r}_{s,a} - \bar{Q}_{s,a}+ \gamma\sum_{s'}\bar{P}_\delta(s'|s,a) \bar{Q}_{s',\bar{\pi}_{s'}})^2.$$
Thus, we can reformulate the problem in~\eqref{equ:formulation_penalty} as the following bilevel problem
\begin{equation}
    \label{equ:bi_level_form}
    \begin{aligned}
\mathop{\text{minimize}}_{\bar{Q},\delta}\text{ } & F(\delta,\bar{Q}, \bar r(\delta, \bar Q)), \\
        \text{subject~to} \text{ } & \bar{r}(\delta, \bar Q) \in \mathop{\arg\min}_{\bar{r}}g(\bar{r},\delta, \bar{Q}),
        \end{aligned}
\end{equation}
where \begin{equation}\label{equ:upper_obj}        F(\delta,\bar{Q},\bar r(\delta, \bar Q)) := \frac{1}{2} \sum_{s,a} ( \bar{r}_{s,a}(\delta, \bar{Q}) -r_{s,a} )^2 +\frac{\rho_\delta}{2}\sum_{s,a}\delta_{s,a}^2+ \frac{\rho_\Phi}{2}\sum_{s,a\neq\bar \pi_s} (\Phi(\bar{Q}_{s,a}+\epsilon-\bar{Q}_{s,\bar\pi_s}))^2
\end{equation}
and the solution to the lower-level problem is denoted as $\bar{r}(\delta, \bar{Q})$, reflecting its dependency on $\delta$ and $\bar Q$.
In~\eqref{equ:bi_level_form}, the lower-level problem is guaranteed to have a unique solution. 
Most common methods for solving bilevel optimization problems with a unique lower-level solution is based on the implicit function theorem(\cite{hong2023two,ghadimi2018approximation}). 
 It is important to note that we only designate $\bar r$ as the lower-level variable in this formulation. 
Setting $\bar{Q}$ 
as a lower-level variable could introduce a double-sampling issue, as discussed in the penalty reformulation. In addition, 
$\delta$ is component in a simplex set, 
which imposes boundedness constraints and hinders the transformation
 from~\eqref{equ:formulation_penalty} to~\eqref{equ:bi_level_form} from being an equivalent transformation. \par

%\jh{We hereby illustrate the rationale behind the proposed reformulation. The equality constraints in~\eqref{equ:formulation_penalty} are first transformed  into the lower-level objective function, denoted as $g(\bar{r},\delta,\bar{Q})$. To ensure the lower-level problem admits a unique solution, only one of the variables $\bar{r},\delta,$ or $\bar{Q}$ can be cast as the lower-level variable. The issue arises since the number of the equality constraints (i.e., $|\mathcal{S}||\mathcal{A}|$) is less than the cardinality of any combined variables (which is at least $2|\mathcal{S}||\mathcal{A}|$). However, neither $\delta$ nor $\bar{Q}$ is appropriate. Assigning $\bar{Q}$ as the lower-level variable introduces the double-sampling issue as discussed penalty reformulation, while selecting $\delta$ may compromises the equivalence of problem~\eqref{equ:bi_level_form} with~\eqref{equ:formulation_penalty} due to its limited representation capability on poisoned transition probabilities. Based on above considerations, the reformulated optimization problem~\eqref{equ:bi_level_form} is derived. } \par

%\jh{Considering the dependency of the lower-level problem's solution with $\delta$ and $\bar Q$, we denote the solution as $\bar{r}(\delta, \bar{Q})$, which has an explicit form $\bar{r}_{s,a}(\delta, \bar{Q})=\bar{Q}_{s,a}-\gamma\sum_{s'}\bar{P}_\delta(s'|s,a)\bar{Q}_{s',\bar{\pi}_{s'}}$, as indicated 
%in~\eqref{equ:formulation_penalty}.
Lastly, we will address the gradient computations necessary for resolving the bilevel optimization problem. 
Computing the gradient of $F(\delta,\bar{Q}, \bar r(\delta, \bar Q))$
 with respect to $\delta$ or $\bar{Q}$ involves the terms $\nabla_\delta \bar{r}_{s,a}(\delta, \bar{Q})$ and $\nabla_{\bar{Q}}\bar{r}_{s,a}(\delta, \bar{Q})$. Given the explicit form for  $\bar{r}_{s,a}(\delta, \bar{Q})$:
 $\bar{r}_{s,a}(\delta, \bar{Q})=\bar{Q}_{s,a}-\gamma\sum_{s'}\bar{P}_\delta(s'|s,a)\bar{Q}_{s',\bar{\pi}_{s'}},$ as indicated 
in~\eqref{equ:formulation_penalty}, we derive that, for any $\tilde s,s\in\mathcal{S}, \tilde a,a\in\mathcal{A}$,
    \begin{align}\label{equ:lower_gradient_upper_delta}
            \nabla_{\delta_{\tilde s,\tilde a}}\bar{r}_{s,a}(\delta, \bar{Q}) =&  -\mathbf{1}(\tilde s=s,\tilde a=a)\gamma\sum_{s'}\Big(P(s'|s,a) - \frac{\mathbf{1}(s'\in\mathcal{S}_{s,a}')}{|\mathcal{S}_{s,a}'|}\Big)\bar{Q}_{s',\bar \pi_{s'}},\\
   % \end{equation}          
     % \begin{equation}\label{equ:lower_gradient_upper_Q}
     \nabla_{\bar{Q}_{\tilde s,\tilde a}}\bar{r}_{s,a}(\delta, \bar{Q}) =&  \mathbf{1}(\tilde s=s,\tilde a=a)-\mathbf{1}(\tilde a=\bar \pi_{\tilde s})\gamma \bar{P}_\delta(x|s,a).\label{equ:lower_gradient_upper_Q}
    \end{align}
However,~\eqref{equ:lower_gradient_upper_delta} and~\eqref{equ:lower_gradient_upper_Q}  cannot be directly computed due to unknown $\bar{P}_\delta(s'|s,a)$. Thus, we need to explore methods to approximate these terms, which will be discussed in the next subsection.\par

\subsection{Single-loop Stochastic Gradient Descent for Bilevel Optimization}
\label{subsec:algorithm}
To solve the bilevel optimization problem, a stochastic single-loop gradient descent algorithm for bilevel optimization (\cite{hong2023two}) is employed. The approach constructs a conjugate upper-level gradient using the current lower-level variable. The subsequent analysis unfolds by first computing the exact gradients and then deriving the conjugate upper-level gradient. The stochastic gradients are further obtained by substituting the transition probabilities using sampled transitioned states. \par

We first derive the exact gradients of both the upper- and lower-level variables. The gradient of lower-level objective function $g(\bar{r}, \delta, \bar{Q})$ with respect to $\bar{r}$ is computed as follows: for arbitrary state-action pair $s,a$,
\begin{equation}\label{equ:gradient_bar_r}
        \nabla_{\bar{r}_{s,a}} g(\bar{r}, \delta, \bar{Q}) = \bar{r}_{s,a} - \bar{Q}_{s,a} + \gamma\sum_{s'}\bar{P}_\delta(s'|s,a) \bar{Q}_{s', \bar \pi_{s'}}.
\end{equation}

\begin{algorithm}[t]
    \caption{Online environment poisoning algorithm}
    \label{alg:pseudo-code}
    \textbf{Input}: Initial poisoning strategy $\bar{r}^0, \delta^0$. Value gap $\epsilon$. Step sizes $\alpha_k$'s, $\beta_k$'s, $\lambda_k$'s, weight parameter $\rho_\delta$, and gradually increasing penalty coefficients $\rho_{\Phi, k}$'s. 
    \begin{algorithmic}[1]
        \WHILE{(Agent updates RL policy)}
            \STATE (\textbf{Agent-environment interaction:} Agent interacts with environment using policy $\pi$. The environment further generates batch of transition data $\{\langle s_i,a_i,r_i,s'_i \rangle\}_{i\in I_k}$)
            \STATE \textbf{Attacker poisoning:} Attacker manipulates data $\langle s_i,a_i,r_i,s_i' \rangle_{i\in I_k}$ to poisoned data $\langle s_i,a_i,\bar{r}_{s_i,a_i}, \bar{s}_i' \rangle$ using poisoning strategy $\bar{r}^k$ and $\delta_k$, as described in Section~\ref{subsec:attack_scheme} for transitioned state manipulation. 
            \STATE (\textbf{RL training}: Agent updates RL policy based on poisoned data.)            
           % \noindent
            \STATE  \textbf{Poisoned reward update:} Attacker updates $\bar{r}^{k+1}$ based on $I_k$: for each $i$,
            \begin{equation*}
                \bar{r}^{k+1}_{s_i,a_i} = \bar{r}^k_{s_i,a_i} - \alpha_k (\bar{r}^k_{s_i,a_i} - \bar{Q}^k_{s_i,a_i} + \gamma \bar{Q}^k_{\bar{s}_i', \bar \pi_{\bar{s}_i'}})
            \end{equation*}
                
            %\noindent 
            \STATE \textbf{Poisoned $Q$-value update:} Attacker updates $\bar{Q}^{k+1}$ based on $I_k$: for each $i$,
            \begin{align*}
                \bar{Q}^{k+1}_{s_i,a_i} = & \bar{Q}^k_{s_i,a_i} - \beta_k \Big( \bar{r}^{k+1}_{s_i,a_i} - r_i + \mathbf{1}(a_i\neq\bar \pi_{s_i}) \rho_{\Phi,k} \Phi(\bar{Q}_{s_i,a_i}^{k}+\epsilon - \bar{Q}_{s_i,\bar \pi_{s_i}}^{k})\nonumber\\
                & - \mathbf{1}(a_i=\bar \pi_{s_i})  \rho_{\Phi,k} \sum_{u\neq a_i} \Phi(\bar{Q}_{s_i,u}^k+\epsilon-\bar{Q}_{s_i,a_i}^k)\Big),\\
                \bar{Q}^{k+1}_{\bar{s}_i',\bar \pi_{\bar{s}_i'}} = & \bar{Q}^k_{\bar{s}_i',\bar \pi_{\bar{s}_i'}} + \beta_k\gamma( \bar{r}^{k+1}_{s_i,a_i} - r_i ).
            \end{align*}

            \STATE \textbf{Poisoned transition update:} Attacker uniformly samples $s_i^u$ from $\mathcal{S}_{s_i,a_i}'$, and then updates $\delta^{k+1}$ based on $I_k$: for each $i$,
            \begin{align*}
                \delta_{s_i,a_i}^{k+1} =& \textbf{Proj}_{[0,1]}\left(\delta_{s_i,a_i}^k - \lambda_k \left( \gamma(\bar{r}_{s_i,a_i}^{k+1}-r_i)\bar{Q}^{k+1}_{s_i',\bar \pi_{s_i'}} + \rho_\delta\delta_{s_i,a_i}  \right.\right.\\
                &\left.\left. - \gamma (\bar{r}^{k+1}_{s_i,a_i}-r_i)\sum_{s'}\frac{\mathbf{1}(P(s'|\tilde s,\tilde a)>0)}{|\mathcal{S}_{\tilde s,\tilde a}'|} \bar{Q}_{s', \bar{\pi}_{s'}}\right)\right).
            \end{align*}
        \ENDWHILE
    \end{algorithmic}
    % \footnotetext{Agent's behavior is denoted by including with $( )$}
\end{algorithm}
The gradient of outer objective $F$ with respect to $\bar{Q}$ and $\delta$ is computed as follows: for any $\tilde s\in\mathcal{S}, \tilde a\in\mathcal{A}$, 
\begin{align}
\nabla_{\bar{Q}_{\tilde s,\tilde a}}F(\delta, \bar{Q},\bar r(\delta, &\bar Q))
    =  \bar{r}_{\tilde s,\tilde a}(\delta, \bar{Q}) - r_{\tilde s, \tilde a} +\mathbf{1}(\tilde a\neq\bar \pi_{\tilde s}) \rho_\Phi \Phi(\bar{Q}_{\tilde s,\tilde a}+\epsilon - \bar{Q}_{\tilde s,\bar \pi_{\tilde s}})\label{equ:gradient_bar_Q} \\
    & - \mathbf{1}(\tilde a=\bar \pi_{\tilde{s}}) \Big( \gamma \sum_{s,a}\bar{P}_\delta(\tilde s|s,a)( \bar{r}_{s,a}(\delta, \bar{Q}) - r_{s,a} )+ \rho_\Phi\sum_{a\neq \tilde a} \Phi(\bar{Q}_{\tilde s,a}+\epsilon-\bar{Q}_{\tilde s,\tilde a}) \Big),\notag\\
     \nabla_{\delta_{\tilde s, \tilde a}}F(\delta, \bar{Q}, \bar r(\delta, &\bar Q))
    % =  (\bar{r}_{\tilde s,\tilde a}(\delta, \bar{Q})-r_{\tilde s,\tilde a})\gamma \sum_{s'}\Big( P(s'|\tilde s,\tilde a) - \frac{\mathbf{1}(P(s'|\tilde s,\tilde a)>0)}{|\mathcal{S}_{\tilde s,\tilde a}'|} \Big)\bar{Q}_{s', \bar{\pi}_{s'}} + \rho_\delta \delta_{\tilde s, \tilde a}\label{equ:gradient_delta}.
    =  (\bar{r}_{\tilde s,\tilde a}(\delta, \bar{Q})-r_{\tilde s,\tilde a})\gamma \sum_{s'} P(s'|\tilde s,\tilde a)\bar{Q}_{s',\bar{\pi}_{s'}}+ \rho_\delta \delta_{\tilde s, \tilde a}\notag \\
    & -(\bar{r}_{\tilde s,\tilde a}(\delta, \bar{Q})-r_{\tilde s,\tilde a})\gamma \sum_{s'}\frac{\mathbf{1}(P(s'|\tilde s,\tilde a)>0)}{|\mathcal{S}_{\tilde s,\tilde a}'|} \bar{Q}_{s', \bar{\pi}_{s'}} \label{equ:gradient_delta}.
\end{align}
We will substitute $\bar{r}_{\tilde s,\tilde a}(\delta, \bar{Q})$ in~\eqref{equ:gradient_bar_Q} and~\eqref{equ:gradient_delta} with the lower-level variable $\bar{r}_k$ at the current iteration $k$ due to unavailability of the exact $\bar{r}_{\tilde s,\tilde a}(\delta, \bar{Q})$ in black-box environments. 
This results in the formation of conjugate gradients.   \par

Unbiased estimates of $\nabla_{\bar{r}_{s,a}}g(\bar{r},\delta, \bar{Q})$ and $\nabla_{\delta_{\tilde s, \tilde a}}F(\delta, \bar{Q}, \bar{r}(\delta, \bar{Q}))$ can be readily obtained from transition samples $\langle s,a,r_{s,a},s' \rangle$ and poisoned transition
random draws. However, the formula~\eqref{equ:gradient_bar_Q} used for calculating $\nabla_{\bar{Q}_{\tilde{s},\tilde{a}}}F(\delta, \bar{Q}, \bar{r}(\delta, \bar{Q}))$ includes an inverse probability $\bar{P}_\delta(\tilde s|s,a)$, which is associated with the state transition to $\tilde{s}$ from state action pair $(s,a)$. 
This probability can be estimated using transition data that includes $\tilde{s}$ as the transitioned state. Thus, for a single piece of poisoned transition data  $\langle s,a,\bar{r}_{s,a},\bar{s}' \rangle$, both $\bar{Q}_{s,a}$ and $\bar{Q}_{\bar{s}',\bar{\pi}_{\bar{s}'}}$ will be updated through stochastic gradient descent, with the update for the latter term incorporating the approximation of the inverse probability.

The whole online poisoning attack implemented by the attacker is outlined in Algorithm \ref{alg:pseudo-code}.

\section{A Numerical Experiment} 
\label{sec:numerical_results}

\begin{figure}[ht!]
\vspace{-5pt}
    \centering
    \includegraphics[width=0.8\linewidth]{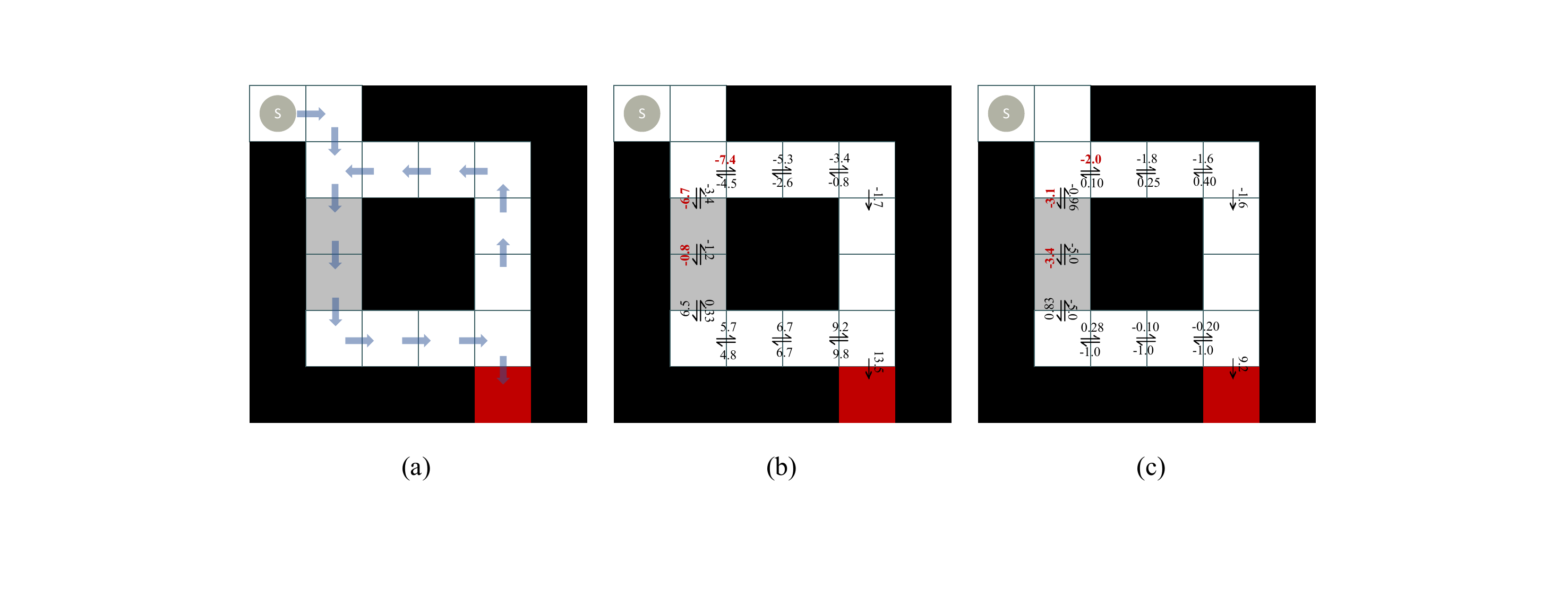}\vspace{-13pt}
    \caption{(a) The target policy implemented by the attacker, which guides the agent to navigate towards the destination while traversing the gray grids; (b) The averaged ultimate $Q$-value learned by the reinforcement learning agent over 5 repeated experiments; (c) The averaged manipulated reward determined by the attacker over 5 repeated experiments. }
    \vspace{-15pt}
    \label{fig:maze_bar_r_Q}
\end{figure}

\textbf{Experimental Setup.}
Consider a maze environment as depicted in Figure \ref{fig:maze_bar_r_Q} (a), where The agent initiates from the grid labeled with `S' and aims to reach the destination (red grid) in the shortest path. We formulate the MDP in the structure $\langle \mathcal{S}, \mathcal{A}, R, P, \gamma \rangle$ as follows. Let $(x,y)$ denote the state of the agent, indicating its position in the $x$th row and $y$th column. The agent is capable of traversing to one of the four adjacent grids. However, we assume that the agent will remain stationary in its current position with 0.7 probability. Each movement incurs a $-1$ reward to encourage finding the shortest path. If the agent traverses the gray grids, which indicate some undesirable states, it is subject to a penalty of $-5$. Conversely, it is awarded with $+10$ if it moves downward to the destination $(6,5)$. The discount factor $\gamma$ is assigned with $0.9$.

At each iteration, the environment generates a transition tuple $\langle s,a,r_{s,a},s' \rangle$ for each $s\in\mathcal{S}$ and $a\in\mathcal{A}$. The agent employs the $Q$-learning algorithm. Note that in the absence of intervention, the agent would navigate to the destination by moving right at grid $(1,1)$. However, as illustrated in Figure \ref{fig:maze_bar_r_Q}(a), the attacker's target policy dictates a downward movement at the same grid location. We set the value gap as $\epsilon=1.0$ and the weight parameter as $\rho_\delta=2.0$. \par

\textbf{Attack Results.}
The $Q$-values for the state-action pairs along the critical path are shown in Figure 2(b), with key values highlighted in red. The agent successfully learned to move downward at grid $(1,1)$ with a value gap of $1.0$, as illustrated in Figure \ref{fig:maze_bar_r_Q}(b). To achieve such effects, the attacker modifies the rewards as illustrated in Figure \ref{fig:maze_bar_r_Q}(c). 

\begin{figure}[t!]
    \centering
    \includegraphics[width=0.97\linewidth]{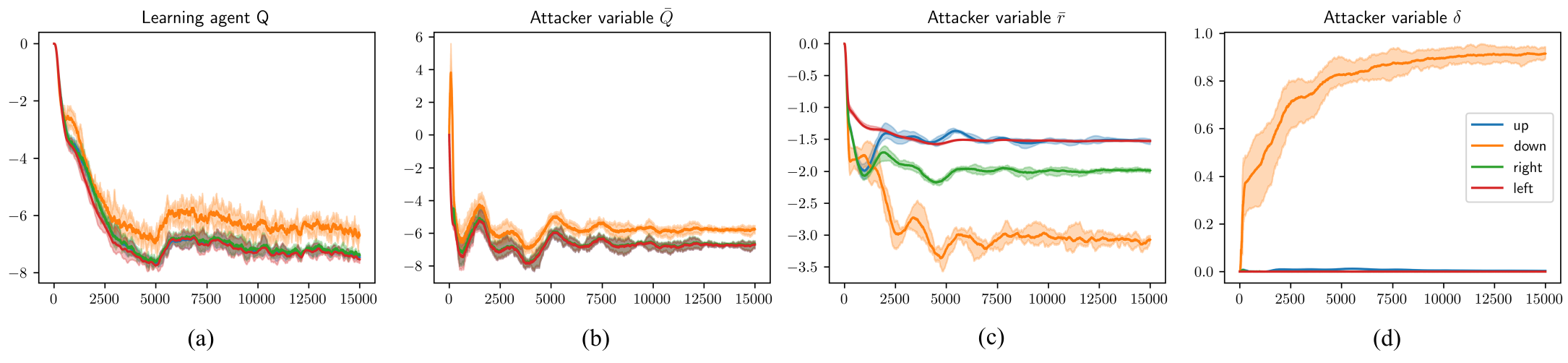}
    \vspace{-16pt}
    \caption{The trajectory of different variables of state $(1,1)$ for four different actions. The experiment is repeated 5 times. The solid line represents the average trajectory, while the shaded area denotes the range between the maximum values and lowest values. }
    \label{fig:state_7}\vspace{-15pt}
\end{figure}

\begin{figure}[t!]
    \centering
    \includegraphics[width=0.92\linewidth]{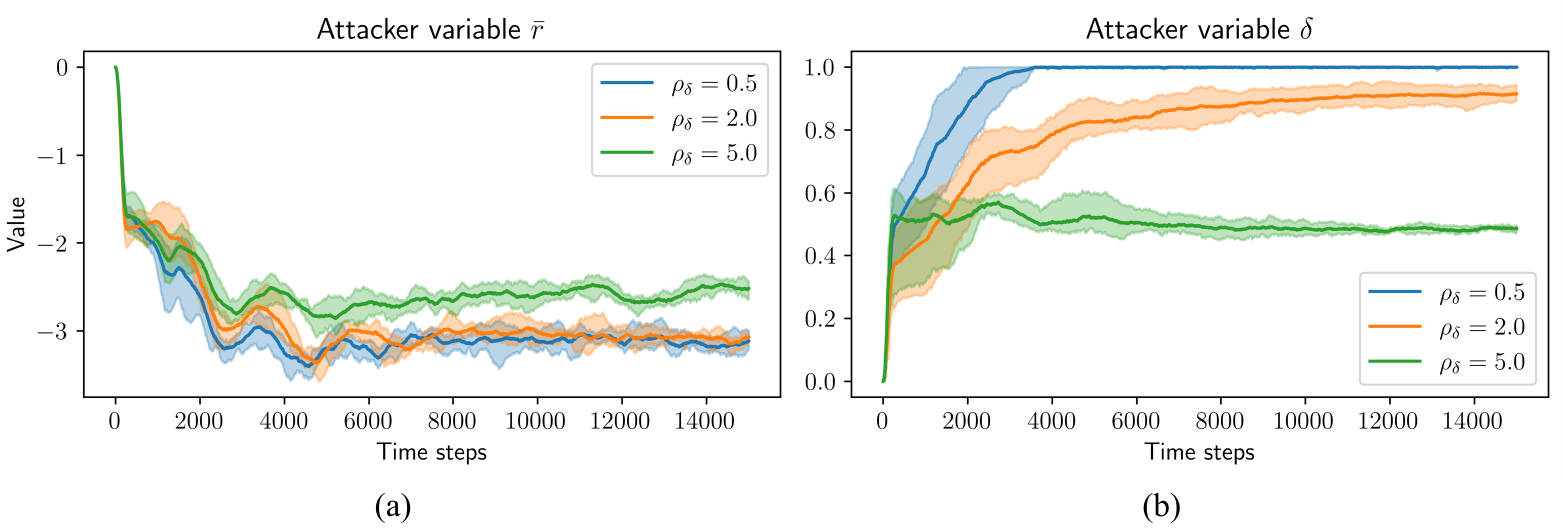}
    \vspace{-13pt}
    \caption{The effect of $\rho_\delta$ on poisoned rewards $\bar{r}$ and transition poisoning intensity $\delta$.}\vspace{-10pt}
    \label{fig:rho_p_repeated}
\end{figure}

Since the state $(1,1)$, which is the intersection of the original path and target path, plays a crucial role in assessing the success of the attack. We specifically plot the trajectories of $\bar{r}, \bar{Q}, \delta$, and agent's $Q$-value at this state in Figure \ref{fig:state_7}. We first observe that the $Q$ value learned by the agent depicted in Figure \ref{fig:state_7}(a) ultimately converges to the attacker's variable $\bar{Q}$ in Figure \ref{fig:state_7}(b), and both exhibit a value gap of $\epsilon=1.0$. To improve the $Q$-value of moving downward, the attacker increases the reward of moving downward and penalizes moving right, as depicted in Figure \ref{fig:state_7}(c). \par

\textbf{Ablation Study for $\rho_\delta$.}
We now evaluate the impact of the weight parameter $\rho_\delta$ on the poisoned rewards $\bar{r}$ and transition poisoning intensity $\delta$. As illustrated in Figure \ref{fig:rho_p_repeated}, increasing $\rho_\delta$ reduces the transition poisoning intensity while amplifying the deviation of the poisoned rewards from their original values. In practice, it can be used to balance the desired level of stealthiness between the poisoned rewards and transition poisoning intensity.
The figure specifically shows the poisoned rewards and transition poisoning intensity at the critical state $(1,1)$ and the key action `move down'.

\section{Conclusion}
We present a novel online attack scheme that is capable of poisoning both the reward function and the transition probabilities in a dynamics-agnostic setting. Our approach employs an optimization-based reformulation that integrates a penalty-based method with a bilevel framework, enabling the derivation of a stochastic gradient. The effectiveness of the proposed method is demonstrated in a maze environment, where the agent's ultimately learned policy aligns with the target malicious policy. The convergence analysis of the proposed algorithm is left for future work.

\bibliography{main}

\begin{thebibliography}{22}
\providecommand{\natexlab}[1]{#1}
\providecommand{\url}[1]{\texttt{#1}}
\expandafter\ifx\csname urlstyle\endcsname\relax
  \providecommand{\doi}[1]{doi: #1}\else
  \providecommand{\doi}{doi: \begingroup \urlstyle{rm}\Url}\fi

\bibitem[Ashcraft and Karra(2021)]{ashcraft2021poisoning}
Chace Ashcraft and Kiran Karra.
\newblock Poisoning deep reinforcement learning agents with in-distribution
  triggers.
\newblock \emph{arXiv preprint arXiv:2106.07798}, 2021.

\bibitem[Chen et~al.(2019)Chen, Li, Li, Jiang, Qi, and
  Song]{chen2019generative}
Xinshi Chen, Shuang Li, Hui Li, Shaohua Jiang, Yuan Qi, and Le~Song.
\newblock Generative adversarial user model for reinforcement learning based
  recommendation system.
\newblock In \emph{International Conference on Machine Learning}, pages
  1052--1061. PMLR, 2019.

\bibitem[Cui et~al.(2024)Cui, Han, Ma, Jiao, and Zhang]{cui2024badrl}
Jing Cui, Yufei Han, Yuzhe Ma, Jianbin Jiao, and Junge Zhang.
\newblock Badrl: Sparse targeted backdoor attack against reinforcement
  learning.
\newblock In \emph{Proceedings of the AAAI Conference on Artificial
  Intelligence}, volume~38, pages 11687--11694, 2024.

\bibitem[Dai et~al.(2018)Dai, Shaw, Li, Xiao, He, Liu, Chen, and
  Song]{dai2018sbeed}
Bo~Dai, Albert Shaw, Lihong Li, Lin Xiao, Niao He, Zhen Liu, Jianshu Chen, and
  Le~Song.
\newblock Sbeed: Convergent reinforcement learning with nonlinear function
  approximation.
\newblock In \emph{International conference on machine learning}, pages
  1125--1134. PMLR, 2018.

\bibitem[Foley et~al.(2022)Foley, Fowl, Goldstein, and
  Taylor]{foley2022execute}
Harrison Foley, Liam Fowl, Tom Goldstein, and Gavin Taylor.
\newblock Execute order 66: targeted data poisoning for reinforcement learning.
\newblock \emph{arXiv preprint arXiv:2201.00762}, 2022.

\bibitem[Ghadimi and Wang(2018)]{ghadimi2018approximation}
Saeed Ghadimi and Mengdi Wang.
\newblock Approximation methods for bilevel programming.
\newblock \emph{arXiv preprint arXiv:1802.02246}, 2018.

\bibitem[Gu et~al.(2016)Gu, Holly, Lillicrap, and Levine]{gu2016deep}
Shixiang Gu, Ethan Holly, Timothy~P Lillicrap, and Sergey Levine.
\newblock Deep reinforcement learning for robotic manipulation.
\newblock \emph{arXiv preprint arXiv:1610.00633}, 1:\penalty0 1, 2016.

\bibitem[Hong et~al.(2023)Hong, Wai, Wang, and Yang]{hong2023two}
Mingyi Hong, Hoi-To Wai, Zhaoran Wang, and Zhuoran Yang.
\newblock A two-timescale stochastic algorithm framework for bilevel
  optimization: Complexity analysis and application to actor-critic.
\newblock \emph{SIAM Journal on Optimization}, 33\penalty0 (1):\penalty0
  147--180, 2023.

\bibitem[Kiourti et~al.(2020)Kiourti, Wardega, Jha, and Li]{kiourti2020trojdrl}
Panagiota Kiourti, Kacper Wardega, Susmit Jha, and Wenchao Li.
\newblock Trojdrl: evaluation of backdoor attacks on deep reinforcement
  learning.
\newblock In \emph{2020 57th ACM/IEEE Design Automation Conference (DAC)},
  pages 1--6. IEEE, 2020.

\bibitem[Liu and Lai(2021)]{liu2021provably}
Guanlin Liu and Lifeng Lai.
\newblock Provably efficient black-box action poisoning attacks against
  reinforcement learning.
\newblock \emph{Advances in Neural Information Processing Systems},
  34:\penalty0 12400--12410, 2021.

\bibitem[Liu and Lai(2023)]{liu2023efficient}
Guanlin Liu and Lifeng Lai.
\newblock Efficient adversarial attacks on online multi-agent reinforcement
  learning.
\newblock \emph{Advances in Neural Information Processing Systems},
  36:\penalty0 24401--24433, 2023.

\bibitem[Ma et~al.(2019)Ma, Zhang, Sun, and Zhu]{ma2019policy}
Yuzhe Ma, Xuezhou Zhang, Wen Sun, and Jerry Zhu.
\newblock Policy poisoning in batch reinforcement learning and control.
\newblock \emph{Advances in Neural Information Processing Systems}, 32, 2019.

\bibitem[Nedi{\'c} and Tatarenko(2020)]{nedic2020convergence}
Angelia Nedi{\'c} and Tatiana Tatarenko.
\newblock Convergence rate of a penalty method for strongly convex problems
  with linear constraints.
\newblock In \emph{2020 59th IEEE Conference on Decision and Control (CDC)},
  pages 372--377. IEEE, 2020.

\bibitem[Neff(2016)]{neff2016talking}
Gina Neff.
\newblock Talking to bots: Symbiotic agency and the case of tay.
\newblock \emph{International Journal of Communication}, 2016.

\bibitem[O'Kelly et~al.(2018)O'Kelly, Sinha, Namkoong, Tedrake, and
  Duchi]{o2018scalable}
Matthew O'Kelly, Aman Sinha, Hongseok Namkoong, Russ Tedrake, and John~C Duchi.
\newblock Scalable end-to-end autonomous vehicle testing via rare-event
  simulation.
\newblock \emph{Advances in neural information processing systems}, 31, 2018.

\bibitem[Rakhsha et~al.(2020)Rakhsha, Radanovic, Devidze, Zhu, and
  Singla]{rakhsha2020policy}
Amin Rakhsha, Goran Radanovic, Rati Devidze, Xiaojin Zhu, and Adish Singla.
\newblock Policy teaching via environment poisoning: Training-time adversarial
  attacks against reinforcement learning.
\newblock In \emph{International Conference on Machine Learning}, pages
  7974--7984. PMLR, 2020.

\bibitem[Rakhsha et~al.(2021)Rakhsha, Zhang, Zhu, and
  Singla]{rakhsha2021reward}
Amin Rakhsha, Xuezhou Zhang, Xiaojin Zhu, and Adish Singla.
\newblock Reward poisoning in reinforcement learning: Attacks against unknown
  learners in unknown environments.
\newblock \emph{arXiv preprint arXiv:2102.08492}, 2021.

\bibitem[Rangi et~al.(2022)Rangi, Xu, Tran-Thanh, and
  Franceschetti]{rangi2022understanding}
Anshuka Rangi, Haifeng Xu, Long Tran-Thanh, and Massimo Franceschetti.
\newblock Understanding the limits of poisoning attacks in episodic
  reinforcement learning.
\newblock \emph{arXiv preprint arXiv:2208.13663}, 2022.

\bibitem[Wu et~al.(2023)Wu, McMahan, Zhu, and Xie]{wu2023reward}
Young Wu, Jeremy McMahan, Xiaojin Zhu, and Qiaomin Xie.
\newblock Reward poisoning attacks on offline multi-agent reinforcement
  learning.
\newblock In \emph{Proceedings of the aaai conference on artificial
  intelligence}, volume~37, pages 10426--10434, 2023.

\bibitem[Xu et~al.(2022)Xu, Qu, and Rabinovich]{xu2022spiking}
Hang Xu, Xinghua Qu, and Zinovi Rabinovich.
\newblock Spiking pitch black: Poisoning an unknown environment to attack
  unknown reinforcement learners.
\newblock In \emph{Proceedings of the 21st International Conference on
  Autonomous Agents and Multiagent Systems}, pages 1409--1417, 2022.

\bibitem[Xu and Singh(2023)]{xu2023black}
Yinglun Xu and Gagandeep Singh.
\newblock Black-box targeted reward poisoning attack against online deep
  reinforcement learning.
\newblock \emph{arXiv preprint arXiv:2305.10681}, 2023.

\bibitem[Zhang et~al.(2020)Zhang, Ma, Singla, and Zhu]{zhang2020adaptive}
Xuezhou Zhang, Yuzhe Ma, Adish Singla, and Xiaojin Zhu.
\newblock Adaptive reward-poisoning attacks against reinforcement learning.
\newblock In \emph{International Conference on Machine Learning}, pages
  11225--11234. PMLR, 2020.

\end{thebibliography}

\end{document}